\def\year{2021}\relax
\documentclass[letterpaper]{article} 
\usepackage{aaai21}  
\usepackage{times}  
\usepackage{helvet} 
\usepackage{courier}  
\usepackage[hyphens]{url}  
\usepackage{graphicx} 
\usepackage{natbib}  
\usepackage{caption} 
\usepackage[super]{nth}
\usepackage{enumitem}

\title{SciDr at SDU-2020 : IDEAS - Identifying and Disambiguating Everyday Acronyms for Scientific Domain}
\author {
    Aadarsh Singh, 
    Priyanshu Kumar \thanks{Authors have equal contribution.}\\ 
}
\affiliations {
    Indian Institute of Technology (Indian School of Mines)\\
    Dhanbad, Jharkhand, India\\
    aadarshsingh191198@gmail.com, kpriyanshu256@gmail.com
}

\begin{document}

\maketitle

\begin{abstract}
We present our systems submitted for the shared tasks of Acronym Identification (AI) and Acronym Disambiguation (AD) held under Workshop on SDU. We mainly experiment with BERT and SciBERT. In addition, we assess the effectiveness of ``BIOless" tagging and blending along with the prowess of ensembling in AI. For AD, we formulate the problem as a span prediction task, experiment with different training techniques and also leverage the use of external data. Our systems rank \nth{11} and \nth{3} in AI and AD tasks respectively.
\end{abstract}

\section{Introduction}


\noindent An acronym is an abbreviation formed from the initial letters of other words and pronounced as a word. The usage of acronyms in articles and speech has increased as it avoids the effort of remembering long complex terms. However, this increased usage of acronyms has also caused new issues of Acronym Identification (AI) and of Acronym Disambiguation (AD). AI is the process of identifying which parts of a sentence constitute the acronyms and their corresponding long forms, whereas AD is the process of correctly predicting the long form expansion of an acronym given a context of its usage. AI and AD are beneficial for applications like question answering \citep{ackermann2020resolution} and definition extraction (\citet{kumar2020explainable}, \citet{singh-etal-2020-dsc}). Since, both AI and AD tasks are benefited with domain knowledge, manual identification and disambiguation of acronyms by domain experts is possible. However, it is tiresome and expensive. Hence, there is a dire need to develop intelligent systems that can mimic the role of domain experts and can help us automate the task of AI and AD. 

In this paper, we present our approach for the shared tasks of Acronym Identification and Acronym Disambiguation held under the workshop of Scientific Document Understanding (SDU). The problem of AI is treated as a sequence tagging problem. For AD, we treat it as a span prediction problem i.e. given a sentence containing an acronym and the possible long forms of that acronym, we aim to extract the span from the possible expansions, which is the most appropriate long form of the acronym as per the context in the sentence. We start the experimentation process for AI with rule based models. The experiments on both tasks are then extended to using Transformers \citep{vaswani2017attention} based architecture, BERT \cite{devlin2018bert} as the backbone of the model; followed by SciBERT \cite{beltagy2019scibert}, which too is a BERT-based model, but is pretrained on text from scientific research papers instead of Wikipedia corpus. In addition, for AD, we experiment with different training procedures, aiming to instill knowledge about various topics into our models.

The rest of the paper is organized as follows : Related works have been discussed in section \ref{sec:rel}, followed by a brief description of the shared task datasets in section \ref{sec:data}. The methodology and experimental settings are covered in sections \ref{sec:methods} and \ref{sec:settings}. Sections \ref{sec:results} and \ref{sec:analysis} contain the results and discussion. Section \ref{sec:conclusion} concludes the paper and also includes scope of future work.

\section{Related Work}
\label{sec:rel}

Initial works on AI incorporate the use of rule-based methods. \citet{park2001hybrid} present rule based methods for finding acronyms in free text. They make use of various patterns, text markers and linguistic cue words to detect acronyms and also their definitions. \citet{schwartz2002simple} make use of the fact that majority acronyms and their long forms are found in close vicinity in sentence, with one of them enclosed between parentheses and thus extract short and long pairs from sentences. They also propose an algorithm for identifying correct long forms.

People have also tried to leverage the use of web-search queries and logs to identify acronym-expansion pairs. A framework for automatic acronym extraction on a large scale was proposed by \citet{jain2007acronym}. They scrape the web for candidate sentences (those containing acronym-expansion pairs) and then identify acronyms-expansion pairs using search query logs and search results. They also try to rank acronym expansions by assigning a score to expansions using various factors. \citet{taneva2013mining} target the problem of finding distinct expansions for an acronym. They make use of query click logs and clustering techniques to extract candidate expansions of acronyms and group them such that each group has a unique meaning. They then assign scores to grouped expansions to find the appropriate expansion.

A comprehensive comparative study between rule-based and machine based methods for identifying and resolving acronyms has been done by \citet{harris2019my}. They collect data from various resources and then experiment with machine based algorithms, crowd-sourcing methods and a game based approach.

\citet{liu2017multi} treat AI as a sequence labelling problem and propose Latent-state Neural Conditional Random Fields model (LNCRF), which are superior to CRFs in handling complex sentences by making use of nonlinear hidden layers. The incorporation of neural networks with CRFs enable learning of better representations from manually created features, which help in better performance.

Many works solve AD task by creating word vectors and then using them to rank the candidates of the acronym with reference to its usage. \citet{mcinnes2011using} correlate acronym disambiguation with word sense disambiguation. They create \nth{2} order vectors of all possible long forms and the acronym with the help of word co-occurrences. The correct long form is then found out using cosine similarity between the vectors. \citet{li2018guess} present an end-to-end pipeline for acronym disambiguation in the domain of enterprise. Due to the lack of mapping of acronym to their long forms, they first use data mining techniques to create a knowledge base. Further, they treat acronym disambiguation as a ranking problem and create ranking models using some manually created features. 

With the advent of deep learning, researchers have tried to create more informative word vectors for the previous approach. \citet{wu2015clinical} first use deep learning to create neural word embedding from medical domain data. They combine the word embeddings of a sample text in different ways and then train a Support Vector Machine (SVM) classifier for each acronym. \citet{charbonnier2018using} explore acronym disambiguation in the scientific research domain. They obtain word vectors from text of scientific research papers and create vector representations for the context of the acronym. Distance minimisation between vector of context and acronym expansion, gives the appropriate expansion.

\citet{ciosici2019unsupervised} present an unsupervised approach for acronym disambiguation by treating it as a word prediction problem. They use word2vec \citep{mikolov2013efficient} to simultaneously learn word embeddings and by learning to predict the correct special token (concatenation of short and long form) of a sentence. The obtained word embeddings are used to create representations of the context of the short form and the best expansion of the short form is obtained from the candidates by minimising distance between representations.

Many works also treat AD as a classification problem. \citet{jin2019deep} explore the usage of contextualised BioELMO word embeddings for acronym disambiguation. They train separate BiLSTM classifiers for each acronym which outputs the appropriate expansion when a text is input. They achieve state of the art performance on PubMed dataset. \citet{li2019neural} propose a novel neural topic attention mechanism to learn better contextualised representations for medical term acronym disambiguation. They compare the performance of LSTMs with ELMo embeddings armed with different types of attention mechanisms.

An overview of the submissions made to the shared tasks of AI and AD has been done by the organizers \cite{veyseh2020acronym}.

\section{Datasets}
\label{sec:data}

\citet{veyseh-et-al-2020-what} provide the shared task participants with a dataset for AI and AD tasks called SciAI and SciAD respectively. SciAI contains 17,506 sentences from research papers, in which the boundaries of acronyms and their long forms are labelled using the BIO format. The tag set consists of \textit{B-short}, \textit{B-long}, \textit{I-short}, \textit{I-long} and \textit{O}, ``short" representing the acronym and ``long" representing the expansion respectively. SciAD contains 62,441 instances covering acronyms used in the scientific domain. The dataset contains the sentence, the acronym and the correct expansion of that acronym as per its usage in the sentence. The dataset also contains a dictionary which is a mapping of the acronyms to candidate long forms. Both datasets are different from the existing datasets for AI and AD as they are larger in size and have instances belonging to scientific domain (majority AI and AD datasets belong to the medical domain).

\section{Methodology}
\label{sec:methods}

\subsection{Models}
Since both the tasks are similar, we try out the following models for both of them and then build upon them:

\begin{itemize}

    \item \textbf{BERT} :  BERT, based on the Transformer architecture consists of multi-attention heads which apply a sequence-to-sequence transformation on the input text sequence. The training objectives of BERT make it unique. The Masked Language Model (MLM) learns to predict a masked token using the left and right context of the text sequence. BERT also learns to predict whether two sentences occur in continuation or not (Next Sentence Prediction). 
    
    \item \textbf{SciBERT} : Allen Institute for Artificial Intelligence (AI2) pretrain the base version of BERT (SciBERT) on scientific text from 1.14 million research papers from Semantic Scholar. Owing to the similarity of the domain of the shared task dataset and SciBERT training corpus, we believe the model will be beneficial for the tasks. We use SciBERT with SciVocab in our experiments.
    
\end{itemize}

\subsection{AI}
\subsubsection{Problem Formulation}
We can easily identify the AI task as a NER (Named Entity Recognition) / BIO tagging task. The tags used in the above methods were short-form and long-form labels of the words in BIO format. One of the interesting experiments that we perform is to make use of ``BIOless" tags. Keeping all factors constant, classifiers ought to work better if the number of classes are less. Tagging is a token classification task. Hence, the tagger should perform better if the number of tags are reduced. The following changes are carried out in the training data to obtain ``BIOless" tags :
\begin{enumerate}
    \item \textit{B-short} and \textit{I-short} tags are changed to \textit{B-short}
    \item \textit{B-long} and \textit{I-long} tags are changed to \textit{B-long}
    \item \textit{O} tags are unchanged.
\end{enumerate}

The models are trained and once the results are obtained, the definition of B, I and O tags viz. beginning, inside and outside, are used to reconstruct the original tags. It is done by changing the first tag in a cluster to \textit{B-short} or \textit{B-long} and the rest of them to \textit{I-short} or \textit{I-long}.

\subsubsection{Models}
We experiment with the following models/variations of the models already mentioned :
\begin{itemize}

    \item \textbf{Conditional Random Fields (CRFs)} :  Considering labelling of sentences with POS (Parts Of Speech) tags, it is highly probable that a NOUN is followed by a VERB. Therefore,  these kinds of task fall under a category which is essentially a combination of classification (classifying a word to one of the POS tags) and graphical modelling (one word influences the POS tag of other words). Thus, these tasks involve predicting a large number of variables that depend on each other as well as on other observed variables.
    
    CRFs are a popular probabilistic method suitable for tasks such as this. They combine the ability of graphical models to compactly model multivariate data with the ability of classification methods to perform prediction using large sets of input features. For the current data, we use the following features as input:
    
        For the \textbf{current word} - 
        \begin{enumerate}[label=\alph*.]
            \item The lower cased version of the word
            \item The last three letters of the word
            \item If all characters of the word are upper case
            \item If the word is title cased
            \item The POS tag of the word
            \item The first two characters of the POS tag of the word
            \item If 60\% of the word is uppercase
        \end{enumerate}
        
        For \textbf{neighbouring words} - 
    
        \begin{enumerate}[label=\alph*.]
            \item The lower cased version of the word
            \item If the word is title cased
            \item If all characters of the word are upper case
            \item The POS tag of the word
            \item The first two characters of the POS tag of the word
        \end{enumerate}
        
    \item \textbf{BERT base cased} : We use the cased base version of BERT as the backbone of our Transformer-CRF architecture
    
    \item \textbf{SciBERT cased} : We use the cased version of SciBERT as the backbone of our Transformer-CRF architecture.

\end{itemize}

    \begin{figure*}[htbp]
        \centering
        \includegraphics[width=\textwidth]{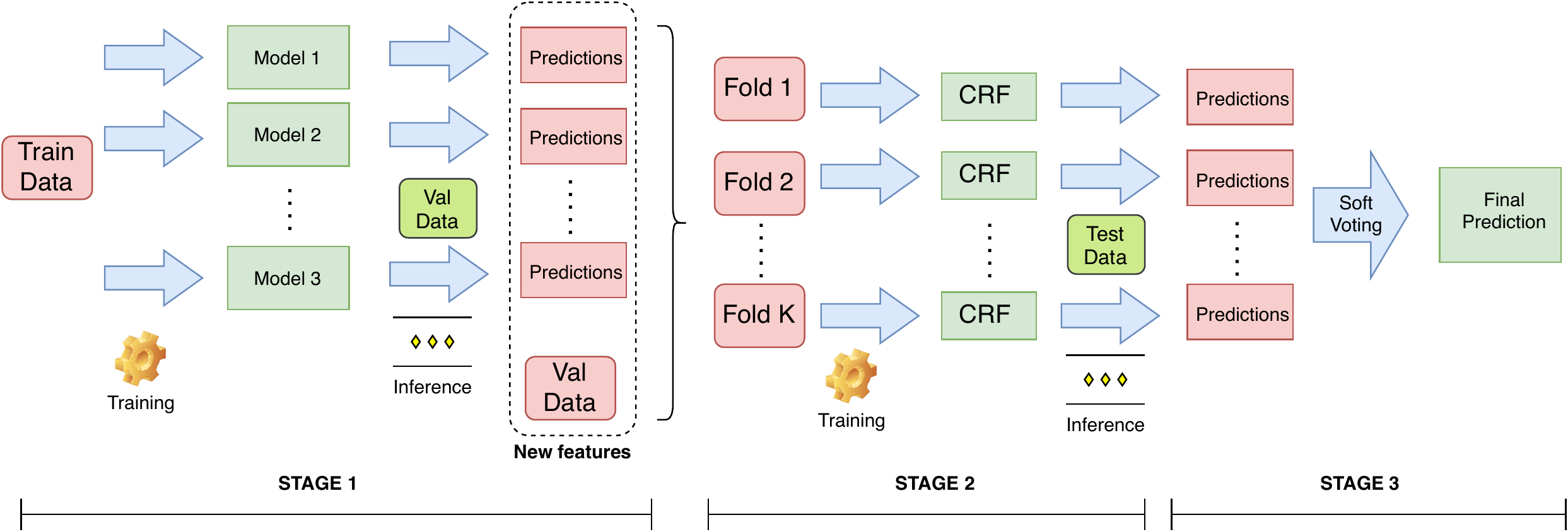} 
        \caption{Blending for AI}
        \label{fig:blend}
    \end{figure*}

\subsubsection{Post Modelling Experiments}
The process of ensembling helped to get a major boost in the score of the base models. We used two kinds of ensembling process:

\begin{itemize}
  
    \item \textbf{Majority Voting/Hard Voting} \citep{wu2006using}: The idea here is to simply go with what the majority of the models in the ensemble method are predicting. In the case of classification, the final prediction is the mode of the predictions of the participating models; similarly, in a tagging task or rather token classification, the final prediction for a given sequence is the sequence of modes of the prediction sequence of the participating models.
    
    Assume $y$ is label, $x$ is the token, $N$ is the total number of base taggers employed and $T_i$ is a function that returns 1 if the prediction of the $i^{th}$ tagger is $y$, otherwise 0.

    Then, $W(y,x)$ is said to be the \textbf{score} and is defined as:
    \[
          W(y,x) = \sum_{i=0}^{N} T_i(y,x)
    \]
    The $y$ with the highest score is chosen as the label of $x$.

    \item \textbf{Blending} \citep{sikdar2017feature} : Hereby, we depict our process of blending models (Figure \ref{fig:blend}). The whole process consists of the following 3 stages:
    \begin{enumerate}[label=\alph*.]
        \item The base models are trained on the training data and then predictions are made on the validation data using these.
        \item The predictions obtained in the previous stage are used as the features for this stage. A CRF is fit on these features using 5-fold cross validation.
        \item The five trained models obtained in the previous stage are then ensembled using majority voting to make the final prediction.
    \end{enumerate}
    
\end{itemize}

\subsection{AD}

\subsubsection{Problem Formulation}

Many existing works on AD solve the problem as a text classification problem, i.e. given a text and an acronym, classify the long form of the acronym or by developing rich word vector representation to extract the most suitable full form out of some candidate long forms. We, instead, treat AD as a span prediction problem. The model predicts the span containing the correct long form from the concatenated text consisting of the acronym, the candidate long forms of that acronym and the sentence (in the same order). The predicted span is then compared with the candidate long forms and the best match is chosen as per Jaccard score. 

Each approach has its own shortcomings. For the classification approach, the size of the model increases with the increase in dictionary size; training models for a large number of classes is difficult. A solution to this problem is to build individual models for acronyms, but the solution might not be feasible if there are many acronyms. For the vector based methods, achieving rich representations is difficult. As for the span prediction approach, the handling of long inputs is difficult and time-consuming. We may have to compromise on the context of the acronym in order to adjust for long sequences.

To prepare our input text for the model, we take advantage of the fact that BERT can encode a pair of sequences together. Therefore, the first sequence is the acronym concatenated with all possible expansions from the dictionary and the second sequence is the input text. Since, some of the input sentences are quite long, we sample tokens from the sentences. In order to input sufficient context of the acronym into the models, we take $n/2$ space delimited tokens to the left of the acronym and $n/2$ space delimited tokens to the right of it, where $n$ is a hyperparameter. We find in our experiments that taking $n$ to be sufficiently large gives almost consistent performance. We fix $n$ to 120 in our experiments.

\begin{figure}[htbp]
\centering
\includegraphics[width=\columnwidth]{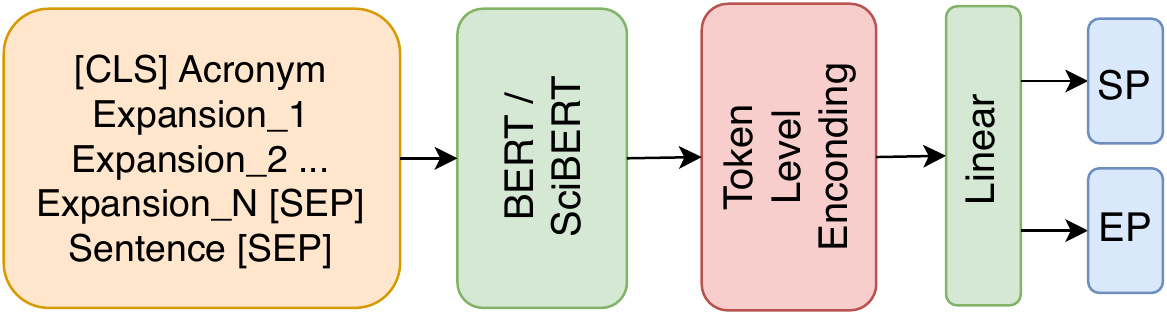} 
\caption{Model Architecture for AD; SP and EP stand for Start Probability and End Probability .}
\label{fig:model}
\end{figure}

We experiment with different training approaches and pretrained weights keeping the architecture of our model constant in all cases. The backbone of the architecture is the  base version of BERT. The sequence outputs of the last layer of BERT (shape = $(batch\_size, max\_len, 768) )$ is passed through a dense layer to reduce its shape to $(batch\_size, max\_len, 2)$. The output is splitted into 2 parts at the \nth{2} axis to get our token level logits for start position and end position. A pictorial representation of the model can be found in Figure \ref{fig:model}. 

\subsubsection{Models}
We experiment with the following models:

\begin{itemize}

    \item \textbf{BERT base uncased} : We use the uncased base version of BERT as the backbone of our model.
    
    \item \textbf{SciBERT uncased} : We use the uncased version of SciBERT as the backbone of our model.

    \item \textbf{SciBERT uncased with fine tuned LM} : The dataset does not contain samples for all acronym expansions. Hence, models trained only on the provided dataset may suffer when it comes to predicting unseen acronym expansions. We try to instill some knowledge of the acronym expansions in our model by fine tuning the MLM. We scrape Wikipedia for articles (using Wikipedia API \url{https://pypi.org/project/wikipedia/} ) related to the long forms of acronyms present in the dictionary and fine tuned the LM of SciBERT using the data. We then use the new fine tuned model weights for the SciBERT backbone and train it for span prediction.

    \item \textbf{SciBERT uncased with 2 stage training} : We train the model in 2 stages using different data. We prepare our own dataset using the articles scrapped from Wikipedia; occurrences of long forms of acronyms are replaced by the acronym. We first train our model on this data and then on the shared task data. This is a supervised approach to help models learn for acronyms and expansions under represented in the shared task data as compared to the above approach which is unsupervised.

\end{itemize}

\subsubsection{Post Modelling Experiments}
\label{sec:post}

\begin{itemize}
    \item \textbf{Ensemble} : Since our approach outputs start and end probability distribution over the entire sequence of tokens, we cannot average probabilities from models using different tokenizers. Keeping the above fact in mind, we average the probabilities from the two best models (as per CV) i.e. SciBERT uncased and SciBERT uncased with 2 stage training. The appropriate acronym expansion is then extracted with the help of this averaged probability, which provides robustness in our predictions.
    
    \item \textbf{Ensemble with post-processing} : We also devise a post processing that can help us rectify some of the mistakes of our models to some extent. All the post-processing does is that if a candidate expansion of an acronym is present in the sentence and the acronym is enclosed within parenthesis in the sentence, then that candidate expansion is predicted as the expansion of the acronym. The motivation for devising this post-processing is discussed in Section \ref{sec:analysis}.
    
\end{itemize}

\section{Experimental Settings}
\label{sec:settings}
For AI task, there are three kinds of experimental settings:
\begin{enumerate}[label=\alph*.]
    \item The base models were trained on the training data and evaluated on the validation data.
    \item For the better performing base models, we concatenate the training and validation data and perform a 5 fold cross-validation on the concatenated dataset.
    \item For blending, we perform a 5 fold cross-validation on the validation data.
\end{enumerate}
For each one of the above settings, training was done for 20 epochs using early stopping with patience of 10. Model optimisation was done using BertAdam with a learning rate of 1e-3, a batch size of 16 and gradient accumulation batch size of 32.

For the AD task, we concatenate the training and validation data and perform a 5 fold stratified cross-validation on the joined dataset (stratified with respect to acronym). The folds are trained for 5 epochs using early stopping with patience of 2 and tolerance of 1e-3. Model optimisation is done using AdamW \citep{loshchilov2018fixing} with a learning rate of 2e-5 and a batch size of 32.

\section{Results}
\label{sec:results}

\subsection{AI}
\label{subsec:results_ai}
The macro F1 scores of our approaches are listed in Table \ref{table:results_AI}. For, the base models, validation is done using the validation data. Only the promising models, in our case SciBERT models, are taken through the arduous cross validation process.

It should also be noted that the folds for the process of cross validation on the modified blending technique are extracted out of the validation data unlike SciBERT models which are cross validated on the combined data (train + validation), and hence the two CV scores are not comparable. The other observations are enumerated as follows:

\begin{enumerate}[label=\alph*.]
    \item The official baseline, though rulebased, surpasses CRF. 
    \item As expected, SciBERT performs better than BERT. 
    \item As for the BIOless variants:
    \begin{itemize}
        \item CRFs see a considerably big difference (0.026) between the BIOless and BIO variants. The hypothesis that ``the tagger should perform better if the number of tags are reduced" seems to fail here. The present task of AI seems a bit complex for CRFs as they do not even surpass the baseline score of 0.84. Hence, it would only be justifiable to treat CRFs as an exception with respect to the hypothesis.
        \item For all the other models/variations, BIOless is pretty close (a difference of 0.0008 or 0.0002 ) or surpasses the BIO variant(with a relatively larger difference 0.0084 or 0.0013).
    \end{itemize}
    \item Based on the test score, BIOless variants perform better than their corresponding BIO counterparts.
    \item The test score undoubtedly shows the eminence of the modified blending technique.

\end{enumerate}

\begin{table}[htbp]
\centering
\begin{tabular}{|p{3cm}|c|c|c|}
    \hline
    \textbf{Model} & \textbf{Val} & \textbf{CV} & \textbf{Test} \\ \hline
    Baseline & 0.8546 & - & 0.8409 \\ \hline
    CRF & 0.8254 & - & - \\ \hline
    CRF BIOless & 0.7994 & - & - \\ \hline
    BERT cased & 0.9145 & - & - \\ \hline
    BERT cased BIOless& 0.9163 & - & - \\ \hline
    SciBERT cased & \textbf{0.9173} & - & 0.8921 \\ \hline
    SciBERT cased BIOless & 0.9165 & - & 0.9005 \\ \hline
    SciBERT cased & - & \textbf{0.9075} & 0.9023 \\ \hline
    SciBERT cased BIOless & - & 0.9073 & 0.9036 \\ \hline
    Blending with mode ensembling & - & 0.8962 & \textbf{0.9090}\\ \hline
\end{tabular} 
\caption{Results of AI task.}
\label{table:results_AI}
\end{table}

Table \ref{table:compareAI} shows a comparison of our results with the top scoring submissions of AI task.

\begin{table}[htbp]
\centering
\begin{tabular}{|c|c|}
    \hline
    \textbf{User / Team Name} & \textbf{Test Score} \\ \hline
    zdq & \textbf{0.9330} \\ \hline
    qinpersevere & 0.9311 \\ \hline
    Mobius & 0.9281 \\ \hline
    SciDr (Us) & 0.9090 \\ \hline
    
\end{tabular} 
\caption{Comparison of AI results}
\label{table:compareAI}
\end{table}

\subsection{AD}

We tabulate the macro F1 score of the models in the cross-validation and test setting (in Table \ref{table:results}). The performance of SciBERT is superior to BERT owing to the similarity of pretraining corpus and task dataset. We also observe that the performance of SciBERT uncased and SciBERT uncased with 2 stage training is almost similar in both cross-validation and test, with the latter performing a bit better than former, whereas the performance of the one with fine-tuned LM is lower. A possible reason for this observation can be attributed to the difference between the source of the data used for fine tuning (Wikipedia) and the shared task data (scientific papers). The usage of extra data created using Wikipedia is beneficial for the model since it contains samples for some acronyms under-represented in the task dataset.

\begin{table}[htbp]
\centering
\begin{tabular}{|p{3cm}|p{1.2cm}|p{1.2cm}|}
    \hline
    \textbf{Model} & \textbf{CV} & \textbf{Test} \\ \hline
    Baseline & - & 0.6097 \\ \hline
    BERT uncased & 0.7549 & 0.8980 \\ \hline
    SciBERT uncased & 0.8423 & 0.9244 \\ \hline
    SciBERT uncased with fine tuned LM & 0.8278 & 0.9194 \\ \hline
    SciBERT uncased with 2 stage training & \textbf{0.8424} & 0.9292 \\ \hline
    Ensemble & - & 0.9303 \\ \hline
    Ensemble with post-processing & - & \textbf{0.9319} \\ \hline
    
\end{tabular} 
\caption{Results of AD task.}
\label{table:results}
\end{table}

Table \ref{table:compareAD} lists the scores of the top submissions for AD task.

\begin{table}[htbp]
\centering
\begin{tabular}{|c|c|}
    \hline
    \textbf{User / Team Name} & \textbf{Test Score} \\ \hline
    DeepBlueAI & \textbf{0.9405} \\ \hline
    qwzhong & 0.9373 \\ \hline
    SciDr (Us) & 0.9319 \\ \hline
    del2z & 0.9266 \\ \hline
    
\end{tabular} 
\caption{Comparison of AD results}
\label{table:compareAD}
\end{table}

\begin{figure*}[htbp]
\centering
\includegraphics[width=\textwidth]{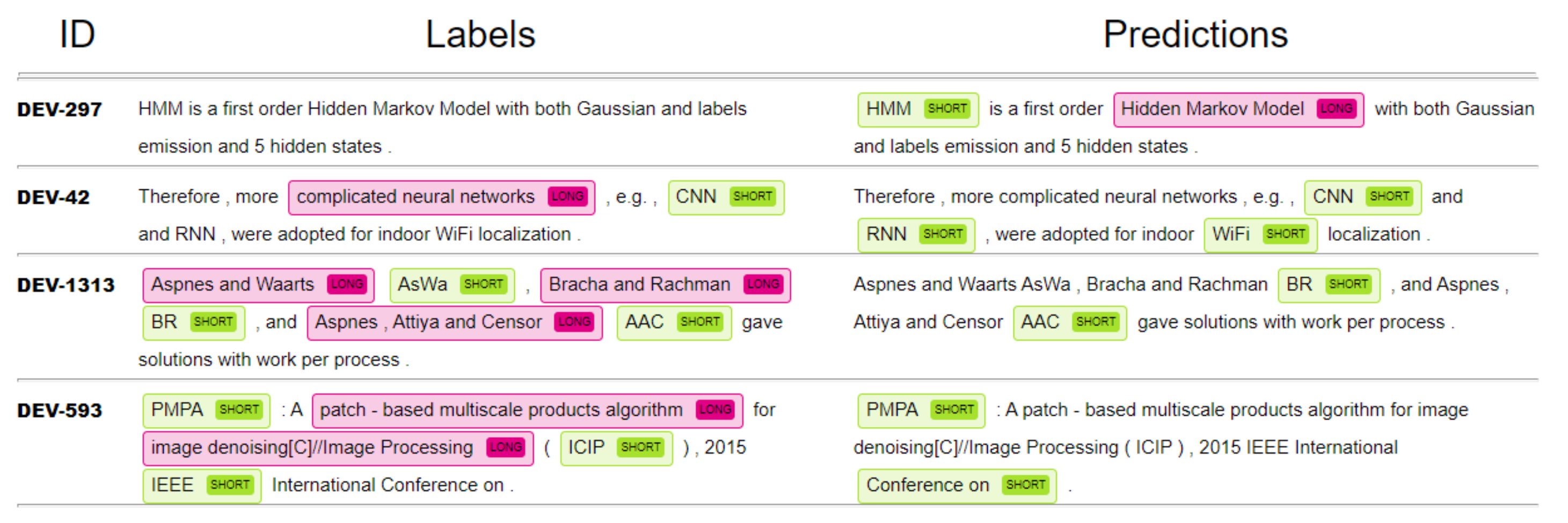} 
\caption{A few erroneously tagged instances for AI.} 
\label{fig:tag_vis}
\end{figure*}

\section{Discussion}
\label{sec:analysis}

\subsection{AI}

The best proposed method for the AI task involves the use of the following three main building blocks:
\begin{itemize}
    \item SciBERT as the base model
    \item BIOless variant
    \item Modified blending technique or the blending method coupled with hard voting.
\end{itemize}

The reason for SciBERT performing better than the BERT model lies in the fact that the pretraining corpus is similar to our dataset. The hypothesis for using BIOless variants instead of the conventional technique seems to hold true (points c, d and e in Subsection \ref{subsec:results_ai}).

\begin{table}[htbp]
\centering
\begin{tabular}{|p{3cm}|c|c|c|}
    \hline
    \textbf{Model} & \textbf{F1} & \textbf{Precision} & \textbf{Recall}\\ \hline
    Baseline & 0.8409 & 0.9131 & 0.7793 \\ \hline
    SciBERT cased BIOless with hard voting & 0.9036 & 0.8987 & 0.9086 \\ \hline
    Blending with mode ensembling & 0.9090 & 0.9097 & 0.9083\\ \hline


\end{tabular} 
\caption{F1, Precision and Recall of some models used in AI Task}
\label{table:detail_score}
\end{table}

Ensembling has always helped in the domain of Machine Learning. The third block viz. modified blending technique is a combination of two propitious methods - blending and hard voting, and ultimately went about to give the best results. The baseline method used by the organizers had a low F1 but the precision obtained was quite good compared to the precision of the SciBERT cased BIOless model with hard voting. The only way to employ the adroitness of the baseline model was to stack it (and some other better performing models) with the SciBERT cased BIOless model. And as is visible in Table \ref{table:detail_score}, the Blended model improved considerably especially with respect to precision.


Figure \ref{fig:tag_vis} represents some of sentences tagged incorrectly by the SciBERT model. Ideally the analysis should have been done on the best model, but it is too complex to interpret it. Having a look at the \textbf{DEV-297} and \textbf{DEV-42}, it is clear that the gold truths have some annotation flaws. HMM is clearly an acronym for Hidden Markov Models and still is not labelled. Similarly, RNN, CNN and WiFi are acronyms for Recurrent Neural Network, Convolutional Neural Network and Wireless Fidelity respectively but only CNN is marked in the ground truth. Also, complicated neural network is no full form but is used to show the complications of RNN and CNN neural networks. Our base model does good in predicting the right tags for there samples.

On the other hand, we find that in \textbf{DEV-1313} and \textbf{DEV-593}, the model has completely failed to identify the long forms, and also misidentified a few short forms. Two likely causes could be as follows:
\begin{itemize}
    \item improper tokenization of the dataset
    \item ``and", ``-", ``of" etc. in between long forms 
\end{itemize}


\begin{table*}[htbp]
\centering
\begin{tabular}{|c|c|p{5cm}|c|c|c|}
    \hline
    \textbf{Id} & \textbf{Acronym} & \textbf{Text} & \textbf{Normal} & \textbf{Stage} & \textbf{Ensemble} \\ \hline
    
    TS-633 & FM & Ultimately , once we select an FM , the ChI becomes a specific operator . & feature map & fuzzy measure & factorization machines \\ \hline
    
    TS-811 & GS & Additionally , using WSE ( GS search ) we obtained 84.4 accuracy with an FPR of 0.157 and AUC value of 0.918 . 	& genetic search & google scholar 's & gold standard \\ \hline
    
    TS-5682 & EL & Thus , with EL system ( ) , only two structures are possible for : ( i ) , and ( ii ) , . & external links & euler - lagrange & entity linking \\ \hline
    
\end{tabular} 
\caption{Mismatch of predictions between SciBERT uncased, SciBERT uncased with 2 stage training and their soft ensemble.}
\label{table:mismatch}
\end{table*}

\subsection{AD}

The formulation of AD as a span prediction problem is quite efficient from the performance and computational expense point of view. A complete cross-validation run under the experimental settings can be performed in 6 hours on an average on a NVIDIA Tesla P100.

Speaking about the results, for the out-of-fold predictions of SciBERT uncased, we observe that the model is incorrect mainly for acronyms which do not have many occurrences in the task dataset. This motivated us to attempt instilling knowledge into our models via external data.

We first examine the differences between the test set predictions of SciBERT uncased, SciBERT uncased with 2 stage training and their ensemble (represented as Normal, Stage and Ensemble respectively) to understand the difference between the models and to find out which model is exhibiting more confidence in its prediction.

\begin{table}[htbp]
\centering
\begin{tabular}{|c|c|p{4cm}|}
    \hline
    \textbf{Id} & \textbf{Acronym} & \textbf{Text} \\ \hline
    TS-5572 & LPP & The LPP can be briefly described as follows . \\ \hline
    TS-5830 & GCN & Effect of both kernels added at end to get actual GCN output . \\ \hline
    
\end{tabular} 
\caption{Instances lacking sufficient context for AD.}
\label{table:context}
\end{table}

We examine those samples where all of the three predictions are different (Table \ref{table:mismatch}). It can be observed that the predictions of SciBERT uncased seem pretty appropriate as per the context and the contributions from the Stage model changes the final prediction. However, there are 92 instances in the test predictions where any of the three predictions are different. These are the instances where the ensemble submission gets the test score boost.

We observe that some of the samples in the test set do not contain sufficient context which can help in acronym disambiguation. This can be an issue and it is difficult to say how the models will behave in such situations. Some of the samples are shown in Table \ref{table:context}. For the text with id \textbf{TS-5572}, the possible long forms of LPP are ``locality preserving projections" and ``load planning problem". Both the models predict one of the expansions and both the expansions seem relevant in the given context. Similar arguments can be given for the text with id \textbf{TS-5830}, where the models get confused between ``global convolution networks" and ``graph convolution networks".

Many of the instances in the test set are such that the long form expansion of the acronym is present in the text and the acronym is present within parentheses. Our models correctly predict the long form for most of these instances, but miss out on a few occasions. This motivated us to devise a post-processing for such instances, where we can directly check for such conditions and predict accordingly, overwriting the model predictions.

\section{Conclusion}
\label{sec:conclusion}

We present our approach for Acronym Identification and Acronym Disambiguation in scientific domain. The usage of SciBERT in both tasks is beneficial because of domain and training corpus similarity. We addressed AI as a tagging problem. Our experiments prove the usefulness of data transformation using BIOless tags, and the adroitness of blending incorporated with hard voting. We approached AD as span prediction problem. Our experimental work demonstrates the effect of pretrained weights, external data, ensembling and post-processing. Our analysis provides some interesting insights into some of the shortcomings of the models and also some of the flaws in the dataset annotation. For future work, we can experiment with data augmentation and observe the behaviour of the models for both AI and AD.

\section{Appendix}

The source code of our approaches for AI and AD can be found at :
\begin{itemize}
    \item AI : \url{https://github.com/aadarshsingh191198/AAAI-21-SDU-shared-task-1-AI}
    \item AD : \url{https://github.com/aadarshsingh191198/AAAI-21-SDU-shared-task-2-AD}
\end{itemize}

\section*{Acknowledgements}

We thank Google Colab and Kaggle for their free computational resources.

\bibliography{aaai21.bib}

\end{document}